\documentclass[10pt,twocolumn,letterpaper]{article}

\usepackage{cvpr}
\usepackage{times}
\usepackage{epsfig}
\usepackage{graphicx}
\usepackage{amsmath}
\usepackage{amssymb}
\usepackage{algorithmic}
\usepackage{algorithm}
\usepackage{subcaption}
\usepackage{multirow, makecell}

\usepackage{xcolor}

\graphicspath{{images/}}

\newcommand{\data}{\mathbb{D}}
\newcommand{\random}{s} 
\newcommand{\canonical}{c} 
\newcommand{\real}{r}  
\newcommand{\gen}{G}
\newcommand{\disc}{D}

\newcommand{\fake}{a}
\newcommand{\image}{x}
\newcommand{\mask}{m}
\newcommand{\depth}{d}

\newcommand{\imagefake}{\image_\fake}
\newcommand{\maskfake}{\mask_\fake}
\newcommand{\depthfake}{\depth_\fake}

\newcommand{\imagecanonical}{\image_\canonical}
\newcommand{\maskcanonical}{\mask_\canonical}
\newcommand{\depthcanonical}{\depth_\canonical}
\newcommand{\imagereal}{\image_\real}
\newcommand{\imagerand}{\image_\random}

\def \qtopt {\textit{QT-Opt}}
\def \webaddress {\url{https://sites.google.com/view/rcan/}}

\newcommand{\bs}{\mathbf{s}}
\newcommand{\ba}{\mathbf{a}}

\newcommand{\states}{\mathcal{S}}
\newcommand{\actions}{\mathcal{A}}

\newcommand{\bellman}{\mathcal{E}}
\newcommand{\E}{\mathbb{E}}

\def \methodname {Randomized-to-Canonical Adaptation Networks}
\def \acronym {\textit{RCAN}}

\makeatletter
\DeclareRobustCommand\onedot{\futurelet\@let@token\@onedot}
\def\@onedot{\ifx\@let@token.\else.\null\fi\xspace}

\def\ie{\emph{i.e}\onedot}

\def\etal{\emph{et al}\onedot}
\makeatother

\usepackage[breaklinks=true,bookmarks=false]{hyperref}

\cvprfinalcopy

\def\cvprPaperID{5847}

\title{Sim-to-Real via Sim-to-Sim:  Data-efficient Robotic Grasping via\ Randomized-to-Canonical Adaptation Networks}

\author{
\textbf{Stephen James$^{1}$, Paul Wohlhart$^2$, Mrinal Kalakrishnan$^2$, Dmitry Kalashnikov$^3$,}\\ \and
\textbf{Alex Irpan$^3$, Julian Ibarz$^3$, Sergey Levine$^{3,5}$, Raia Hadsell$^4$, Konstantinos Bousmalis$^4$} \\
slj12@imperial.ac.uk, \{wohlhart, kalakris\}@x.team,\\
\{dkalashnikov, alexirpan, julianibarz, slevine, raia, konstantinos\}@google.com, 
}

\begin{document}

\maketitle

\begin{abstract}
Real world data, especially in the domain of robotics, is notoriously costly to collect. One way to circumvent this can be to leverage the power of simulation to produce large amounts of labelled data. However, training models on simulated images does not readily transfer to real-world ones. Using domain adaptation methods to cross this ``reality gap'' requires a large amount of unlabelled real-world data, whilst domain randomization alone can waste modeling power. In this paper, we present Randomized-to-Canonical Adaptation Networks (RCANs), a novel approach to crossing the visual reality gap that uses no real-world data. Our method learns to translate randomized rendered images into their equivalent non-randomized, canonical versions. This in turn allows for real images to also be translated into canonical sim images. We demonstrate the effectiveness of this sim-to-real approach by training a vision-based closed-loop grasping reinforcement learning agent in simulation, and then transferring it to the real world to attain 70\% zero-shot grasp success on unseen objects, a result that almost doubles the success of learning the same task directly on domain randomization alone. Additionally, by joint finetuning in the real-world with only 5,000 real-world grasps, our method achieves 91\%, attaining comparable performance to a state-of-the-art system trained with 580,000 real-world grasps, resulting in a reduction of real-world data by more than 99\%.
\end{abstract}

\footnotetext[1]{Imperial College London. Work done while Stephen James was at X}
\footnotetext[2]{X, Mountain View, California, United States}
\footnotetext[3]{Google Brain, United States}
\footnotetext[4]{DeepMind, London}
\footnotetext[5]{University of California Berkeley, Berkeley, California, United States}
\setcounter{footnote}{5}

\section{Introduction}

Deep learning for vision-based robotics tasks is a promising research direction~\cite{sunderhauf2018limits}. However, it necessitates large amounts of real-world data, which is a severe limitation, since real-robot data collection is expensive and cumbersome, often requiring days or even months for a single task~\cite{Levine2016, Pinto2016}. Due to the availability of affordable cloud computing services, it is becoming more attractive to leverage large-scale simulations to collect experience from a large number of agents in parallel. But with this comes the issue of transferring gained experience from simulation to the real world --- a non-trivial task given the usually large domain shift.

\begin{figure}
\includegraphics[width=1.0\linewidth]{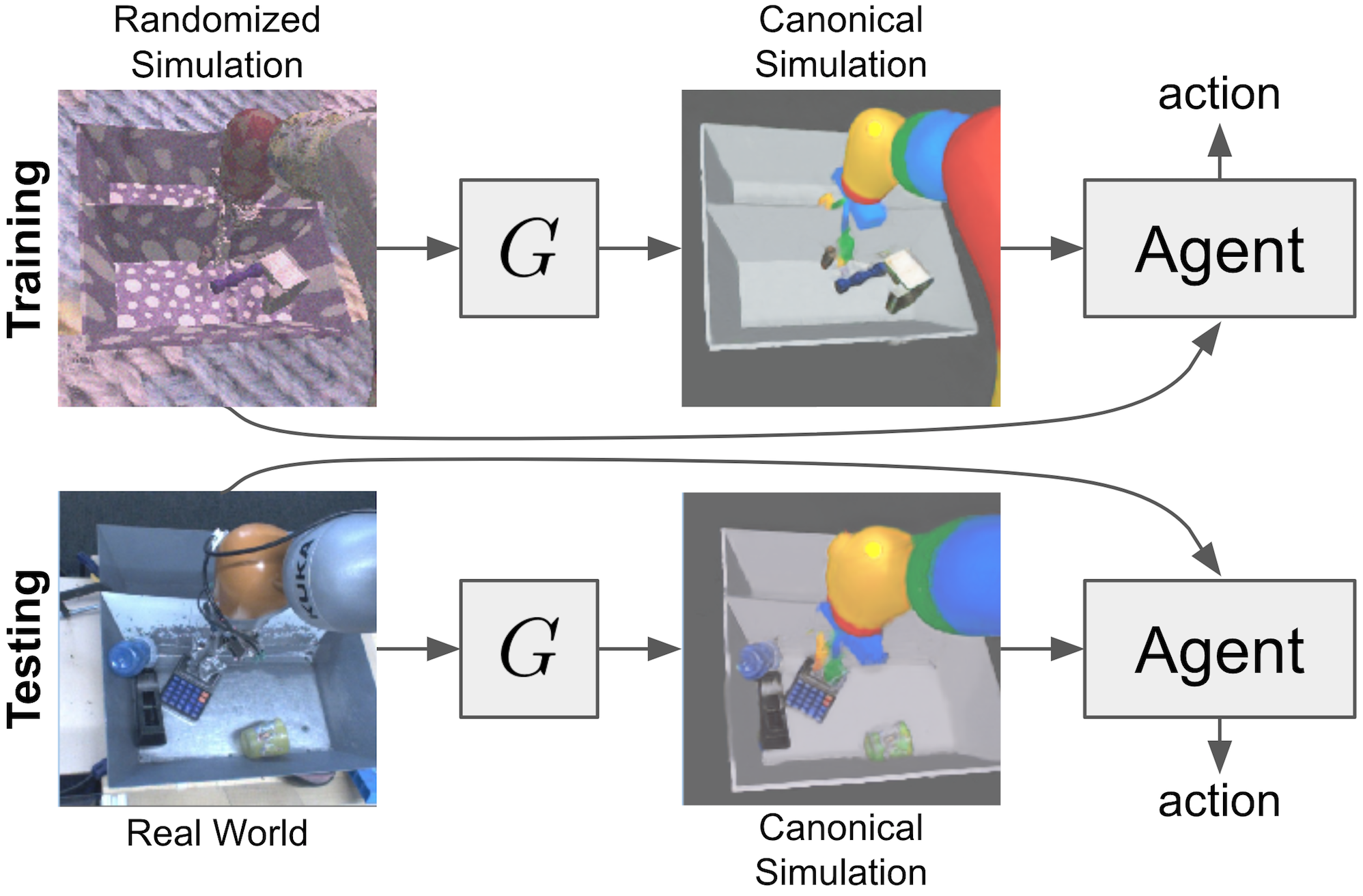}
\caption{We learn a generator that translates randomized simulation images to a chosen canonical simulation version which are then used to train a robot grasping agent (top). The system can then be used to translate real-world images to canonical images, and consequently allow for Sim-to-Real transfer of the agent (bottom). Feeding both source and target images to the agent allows for joint finetuning of the agent in the real world.}
\label{fig:method_summary}
\end{figure}

Reducing the reality gap between simulation and reality is possible with recent advances in visual domain adaptation \cite{ganin2016domain, long2015learning,bousmalis2016domain,shu2018a,bousmalis2017pixelda, yi2017dualgan, zhu2017unpaired, kim2017learning,  shrivastava2017learning, taigman2016unsupervised, hoffman2018cycada}. Such techniques usually require large amounts of unlabelled images from the real world. Although such unlabelled images are easier to capture than labelled, they can still be costly to collect in robotics tasks. Domain randomization~\cite{sadeghi2016cad2rl,tobin2017domain,james2017transferring, matas2018sim, Bousmalis2018, james2018task} is another technique that is particularly popular in robotics, where an agent is trained on a wide range of variations of sensory inputs, with the intention that this forces the input processing layers of the network to extract semantically relevant features in a way that is agnostic to the superficial properties of the image (such as particular textures or particular ways shadows are cast from a constant light source). The intuition is that this leads to a network that extracts the same information from real-world images, featuring yet another variation of the input. However, performing randomization directly on the input of a learning algorithm, as done in related work, makes the task potentially harder than necessary, as the algorithm has to model both the arbitrary changes in the visual domain, while at the same time trying to decipher the dynamics of the task. Moreover, although randomization has been successful in the supervised learning setting, there is evidence that some popular reinforcement learning (RL) algorithms, such as DDPG~\cite{lillicrap2015continuous} and A3C~\cite{mnih2016asynchronous}, can be destabilized by this transfer method~\cite{matas2018sim, Zhang2018a}.

In this paper, we investigate learning vision-based robotic closed-loop grasping, where a robotic arm is tasked with picking up a diverse range of unseen objects, with the help of simulation and the use of as little real-world data as possible. Robotic grasping is an important application area in robotics, but also an exceptionally challenging problem: since a grasping system must successfully pick up previously unseen objects, it is not enough simply to memorize grasps that work well for individual instances, but to generalize and extrapolate from an internal understanding of geometry and physics. This presents a particularly difficult challenge for simulation-to-real-world transfer: besides the distributional shift from simulated images and physics, the system must also handle domain shift in the distribution of objects themselves.

To that end, we propose \textit{\methodname{} (\acronym{})}, a novel approach to crossing the reality gap that translates real-world images into their equivalent simulated versions, but makes use of no real-world data. This is achieved by leveraging domain randomization in a unique way, where we learn to adapt from one heavily randomized scene to an equivalent non-randomized, canonical version. We are then able to train a robotic grasping algorithm in a pre-defined canonical version of our simulator, and then use our \acronym{} model to convert the real-world images to the canonical domain our grasping algorithm was trained on.

Using \acronym{} along with a grasping algorithm that uses \qtopt{}, a recent reinforcement learning algorithm, we achieve almost double the performance in comparison to alternative methods of using randomization. Bootstrapping from this performance, and with the addition of only $5{,}000$ real-world grasps, we are able to achieve higher performance than a system trained with $580{,}000$ real-world grasps. In our particular experiment, none of the objects used during testing are seen during either simulated training or real-world joint finetuning.

Our results also show that \acronym{} (summarized in Figure \ref{fig:method_summary}) is superior to learning a grasping network directly with domain randomization. \acronym{} has additional advantages compared to other simulation-to-real-world transfer methods. Firstly, unlike domain adaptation methods, it does not need any real-world data in order to learn our reality-to-simulation translation function. Secondly, \acronym{} gives an interpretable intermediate output that would otherwise not be available when performing domain randomization directly on the policy. Finally, as our method is trained in a supervised manner and preprocesses the input to the downstream task, it enables the use of RL methods that currently suffer from the stability issues when learning a policy directly from domain randomization~\cite{matas2018sim, Zhang2018a}.

In summary, our contributions are as follows:
\begin{itemize}
\item We present a novel approach of crossing the reality gap by using an image-conditioned generative adversarial network (cGAN)~\cite{isola2017image} to transform randomized simulation images into their non-randomized, canonical versions, which in turn enables real-world images to also be transformed to canonical simulation versions.
\item We show that by using this approach, we are able to train a state-of-the-art vision-based grasping reinforcement learning algorithm (\qtopt{}) purely in simulation and achieve \textbf{70\%} success on the challenging task of grasping previously unseen objects in the real world, almost double the performance obtained by naively using domain randomization on the input of the learning algorithm.
\item We also show that by using \acronym{} and joint finetuning in the real-world with only \textbf{5,000} additional grasping episodes we are able to increase grasping performance to \textbf{91\%}, outperforming \qtopt{} when trained from scratch in the real-world with \textbf{580,000} grasps --- a reduction of over $99\%$ of required real-world samples.
\end{itemize}

\section{Related Work}

\textbf{Robotic grasping} is a well studied problem~\cite{bohg2014data}. Traditionally, grasping was usually solved analytically, where 3D meshes of objects would be used to compute the stability of a grasp against external wrenches~\cite{prattichizzo2008grasping, rodriguez2012caging} or constrain the object's motion~\cite{rodriguez2012caging}. These solutions often assume that the same, or similar objects will be seen during testing, such that point clouds of the test objects can be matched with stored objects based on visual and geometric similarity~\cite{brook2011collaborative, ciocarlie2014towards, hernandez2016team, hinterstoisser2011multimodal, kehoe2013cloud}. Due to this limitation, data-driven methods have become the dominant way to solve grasping~\cite{lenz2015deep, mahler2017dex}. These methods commonly make use of either hand-labeled grasp positions~\cite{lenz2015deep, kappler2015leveraging}, self-supervision~\cite{Pinto2016}, or predicting grasp outcomes~\cite{Levine2016}. State-of-the-art grasping systems typically either operate in an open-loop style, where grasping locations are chosen, and then a motion is executed to complete the grasp~\cite{zeng2018learning, morrison2017cartman, mahler2017dex, ten2017grasp}, or in a closed-loop manner, where grasp prediction is continuously run during motion, either explicitly~\cite{Viereck2017}, or implicitly~\cite{Kalashnikov2018}.

\textbf{Simulation-to-real-world transfer} concerns itself with learning skills in simulation and then transferring them to the real world, which reduces the need for expensive real-data collection. However, it is often not possible to naively transfer such skills directly due to the visual and dynamics differences between the two domains~\cite{james20163d}. Numerous works have looked into enabling such transfer both in computer vision and robotics. In the context of robotic manipulation in particular, Saxena \etal~\cite{saxena2008robotic} used rendered objects to learn a vision-based grasping model. Rusu \etal ~\cite{rusu2016sim} introduced progressive neural networks that help adapt an existing deep reinforcement learning policy trained from pixels in simulation to the real world for a reaching task. Other works have considered simulation-to-real world transfer using only depth images~\cite{viereck2017learning, gualtieri2016high}. Although this may be an attractive option, using depth cameras alone is not suitable for all situations, and coupled with the low cost of simple RGB cameras, there is considerable value in studying transfer in systems that solely use monocular RGB images. Although in this work we use depth estimation from RGB input as an auxiliary task to aid with our randomized-to-canonical image translation model, we neither use depth sensors in the real world, nor do we use our estimated depth during training.

Data augmentation has been a standard tool in computer vision for decades. More recently, and as a way to avoid overfitting, the random application of cropping, flipping samples horizontally, and photometric variations to input images were used to train AlexNet~\cite{krizhevsky2012imagenet} and many more subsequent deep learning models. In robotics, a number of recent works have examined using randomized simulated environments~\cite{tobin2017domain,james2017transferring, matas2018sim, Bousmalis2018, james2018task,sadeghi2018sim2real} specifically for simulation-to-real world transfer for grasping and other similar manipulation tasks, extending on prior work on randomization for collision-free robotic indoor flight~\cite{sadeghi2016cad2rl}. These works apply randomization in the form of random textures, lighting, and camera position, allowing the resulting algorithm to become invariant to domain differences and applicable to the real world. There have been more robotics works that do not use vision, but that apply domain randomization on physical properties of the simulator to aid transferability~\cite{mordatch2015ensemble, rajeswaran2016epopt, antonova2017reinforcement, yu2017preparing, peng2018sim}. Recently, Chebotar \etal ~\cite{chebotar2018closing} have specifically looked into learning, from few real-world trajectories, the optimal distribution of such simulation properties, for transfer of policies learned in simulation to the real world. All of these methods learn a policy directly on randomization, whilst our method instead utilizes domain randomization in a novel way in order to learn a randomized-to-canonical adaption function to gain an interpretable intermediate representation and achieve superior results in comparison to learning directly on randomization.

Visual domain adaptation~\cite{patel2015visual, csurka2017domain} is a process that allows a machine learning model trained with samples from a source domain to generalize to a target domain, by utilizing existing but (mostly) unlabeled target data. In simulation-to-reality transfer, the source domain is usually the simulation, whereas the target is the real world. Prior methods can be split into: \textsl{(1)} feature-level adaptation, where domain-invariant features are learned between source and target domains~\cite{gopalan2011domain, gong2012geodesic,sun2015return,caseiro2015beyond,ganin2016domain, long2015learning,bousmalis2016domain,shu2018a}, or \textsl{(2)} pixel-level adaptation, which focuses on re-stylizing images from the source domain to make them look like images from the target domain~\cite{bousmalis2017pixelda, yi2017dualgan, zhu2017unpaired, kim2017learning,  shrivastava2017learning, taigman2016unsupervised, hoffman2018cycada}. Pixel-level domain adaptation differs from image-to-image translation techniques~\cite{isola2017image,chen2017photographic,yoo2016pixel}, which deal with the easier task of learning such a re-stylization from matching pairs of examples from both domains. Our technique can be seen as an image-to-image translation model that transforms randomized renderings from our simulator to their equivalent non-randomized, canonical ones.

In the context of robotics, visual domain adaptation has also been used for simulation-to-real-world transfer~\cite{tzeng2016adapting, stein2018genesis, Bousmalis2018}. Bousmalis \etal~\cite{Bousmalis2018}, introduced the GraspGAN method, which combines pixel-level with feature-level domain adaptation to limit the amount of real data needed for learning grasping. Although the task is similar to ours, GraspGAN required significant amounts of unlabeled real-world data that were previously collected by a variety of pre-existing grasping networks. Our method can be viewed as orthogonal to existing domain adaptation methods and GraspGAN: the process of training the adapter could make use of unlabeled real-world data by incorporating ideas from domain adaptation in the form of additional auxiliary losses to improve performance further. Although in this work we do explore using our simulation-trained policy to collect labeled real-world data for joint finetuning, the combination with domain adaptation techniques is proposed as a promising future research direction.

The reverse, \ie reality-to-simulation transfer, has been examined recently by Zhang \etal~\cite{Zhang2018a} in the context of a simple robotic driving task. The approach has certain advantages, namely the learning algorithm is trained only in simulation, and during inference the real-world images are adapted to look like simulated ones. This decouples adaptation from training and if the real-world environment changes, it is only the adaptation model that needs to be re-learned. We also explore  reality-to-simulation transfer, but unlike~\cite{Zhang2018a}, which uses CyCaDA~\cite{hoffman2018cycada} and unlabeled real-world data, we do so only in simulation, by learning to adapt randomized images from our simulator to their equivalent non-randomized versions, which allows data-efficient transfer of our model to the real-world.

\section{Background}
We demonstrate our approach by using a recent reinforcement algorithm, \textit{Q-function Targets via Optimization} (\qtopt{})~\cite{Kalashnikov2018}, though our method is compatible with any reinforcement learning or imitation learning algorithm, as we are only adapting the input. \qtopt{} is a state-of-the-art method for vision base grasping, which made it an ideal choice as a baseline for a direct comparison. Below, we will cover the fundamentals of Q-learning and then provide an overview of \qtopt{}.

In reinforcement learning, we assume an agent interacting with an environment consisting of states $\bs \in \states$, actions $\ba \in \actions$, and a reward function $r(\bs_t,\ba_t)$, where $\bs_t$ and $\ba_t$ are the state and action at time step $t$ respectively. The goal of the agent is then to discover a policy that results in maximizing the total expected reward. One way to achieve such a policy is to use the recently proposed \qtopt{}~\cite{Kalashnikov2018} algorithm. \qtopt{} is an off-policy, continuous-action generalization of Q-learning, where the goal is to learn a parametrized Q-function (or state-action value function). This can be learned by minimizing the Bellman error:
\begin{equation}
\bellman(\theta) = \E_{(\bs,\ba,\bs') \sim p(\bs,\ba,\bs')} \left[ D \left(
Q_\theta(\bs,\ba) ,Q_T(\bs,\ba,\bs') \right)
\right],
\label{eq:bellman}
\end{equation}
where $Q_T(\bs,\ba,\bs') = r(\bs,\ba) + \gamma V(\bs')$ is a \emph{target value}, and $D$ is a divergence metric, defined as the cross-entropy function in this case. Much like other works in RL, stability was improved by the introduction of two target networks. The target value $V(\bs')$ was computed via a combination of Polyak averaging and clipped double Q-learning to give $V(\bs') = \min_{i=1,2} Q_{\bar{\theta}_i}(\bs',\arg\max_{\ba'} Q_{\bar{\theta}_1}(\bs',\ba'))$.
\qtopt{} differs from other methods primarily with regards to action selection. Rather than selecting actions based on the argmax: $\pi_{\bar{\theta}_1}(\bs) = \arg\max_{\ba} Q_{\bar{\theta}_1}(\bs,\ba)$, \qtopt{} instead evaluates the argmax via a stochastic optimization algorithm over $\ba$; in this case, the cross-entropy method (CEM)~\cite{rubinstein2004cross}. 

\section{Method}

\begin{figure}
\includegraphics[width=1.0\linewidth]{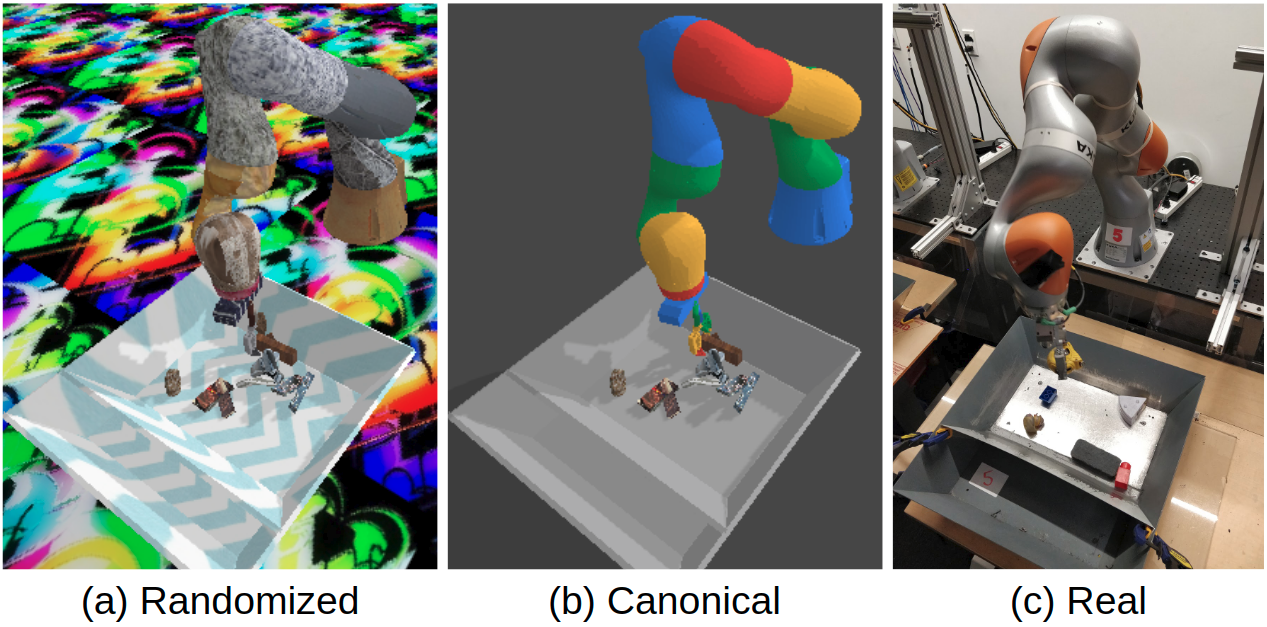}
\caption{The setup used in our approach. A dataset of observations from a randomized version of a simulated environment (a) are paired with observations from a canonical version of the same environment (b) in order to learn an adaptation function and allow observations from the real-world (c) to be transformed into observations looking as if they came from the canonical simulation environment.}
\label{fig:randomized_canonical_real}
\end{figure} 

Our method, \methodname{} (\acronym{}), consists of an image-conditioned generative adversarial network (cGAN)~\cite{isola2017image} that transforms images from  randomized simulated environments (an example is show in Figure \ref{fig:randomized_canonical_real}a) into images that seem similar to those obtained from a non-randomized, canonical one (Figure \ref{fig:randomized_canonical_real}b). Once trained, the cGAN generator is also able to transform real-world images into images that seem as if they were obtained from the canonical simulation environment. We are then able to train a reinforcement learning algorithm (in this case \qtopt{}) fully in simulation, and use such a generator to enable the trained policy to act in the real-world.

The approach assumes 3 domains: the randomized simulation domain, the canonical simulation domain, and the real-world domain. Let $\data = \{(\image_{\random}, \image_{\canonical}, \mask_{\canonical}, \depth_{\canonical})_j\}_{j=1}^{N}$ be a dataset of $N$ training samples, where each sample is a tuple containing an RGB image $\image_\random$ from the randomization (source) domain, an RGB image $\image_\canonical$ from the canonical (target) domain (with semantic content, \ie scene configuration, matching that of $\image_\random$), a segmentation mask $\mask_{\canonical}$, and a depth image $\depth_{\canonical}$. Both the segmentation mask and depth mask are only used as auxiliary tasks during the training of our generator. The \acronym{} generator function $\gen(\image) \rightarrow \{\imagefake, \maskfake, \depthfake\}$, maps an image $\image$ from any domain to an adapted image $\imagefake$, segmentation mask $\maskfake$, and depth image $\depthfake$, such that they appear to belong to the canonical domain. 

\subsection{RCAN Data Generation}
\label{sec:datageneration}

In order to learn this translation $\gen$, we need pairs of observations capturing the robot in interaction with the scene, with one observation showing the scene in its canonical version and the other one showing the same scene but with randomization applied, as shown in image \textit{(a)} and \textit{(b)} of Figure~\ref{fig:randomized_canonical_real}. Our simulated environments are based on the Bullet physics engine and use the default renderer~\cite{coumans2018pybullet}. They are built to roughly correspond to the real word, and include a Kuka IIWA, a tray, an over-the-shoulder camera aimed at the tray, and a set of graspable objects. Graspable objects consist of a combination of 1,000 procedurally generated objects (consisting of randomly merged geometric shapes), and 51,300 realistic objects from 55 categories obtained from the ShapeNet repository~\cite{shapenet}. 

\begin{figure*}
\centering
\begin{subfigure}[b]{0.49\linewidth}
\includegraphics[width=\textwidth]{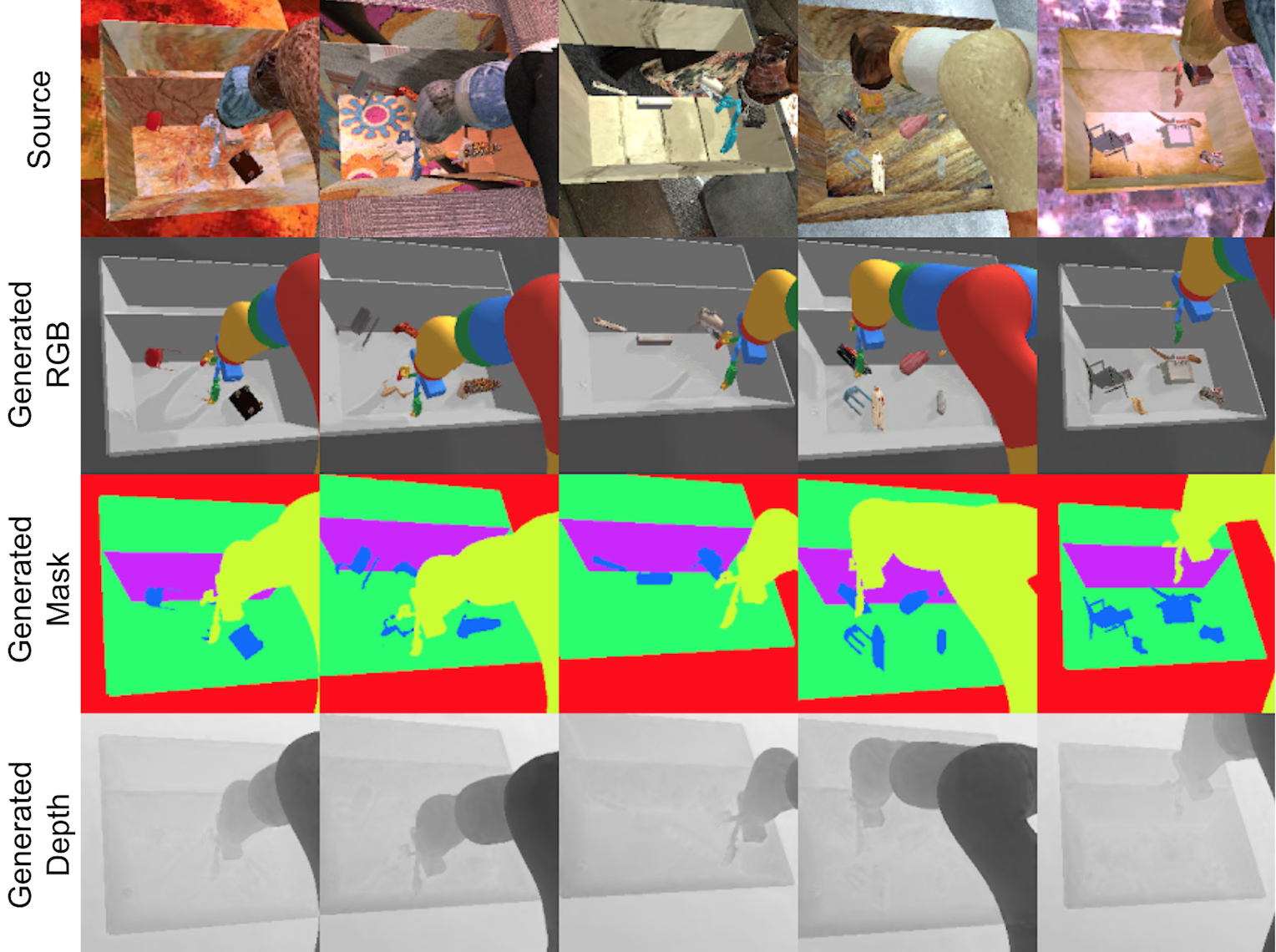}
\caption{Randomized-to-canonical samples.}
\label{subfig:simtosim}
\end{subfigure}
\begin{subfigure}[b]{0.49\linewidth}
\includegraphics[width=\linewidth]{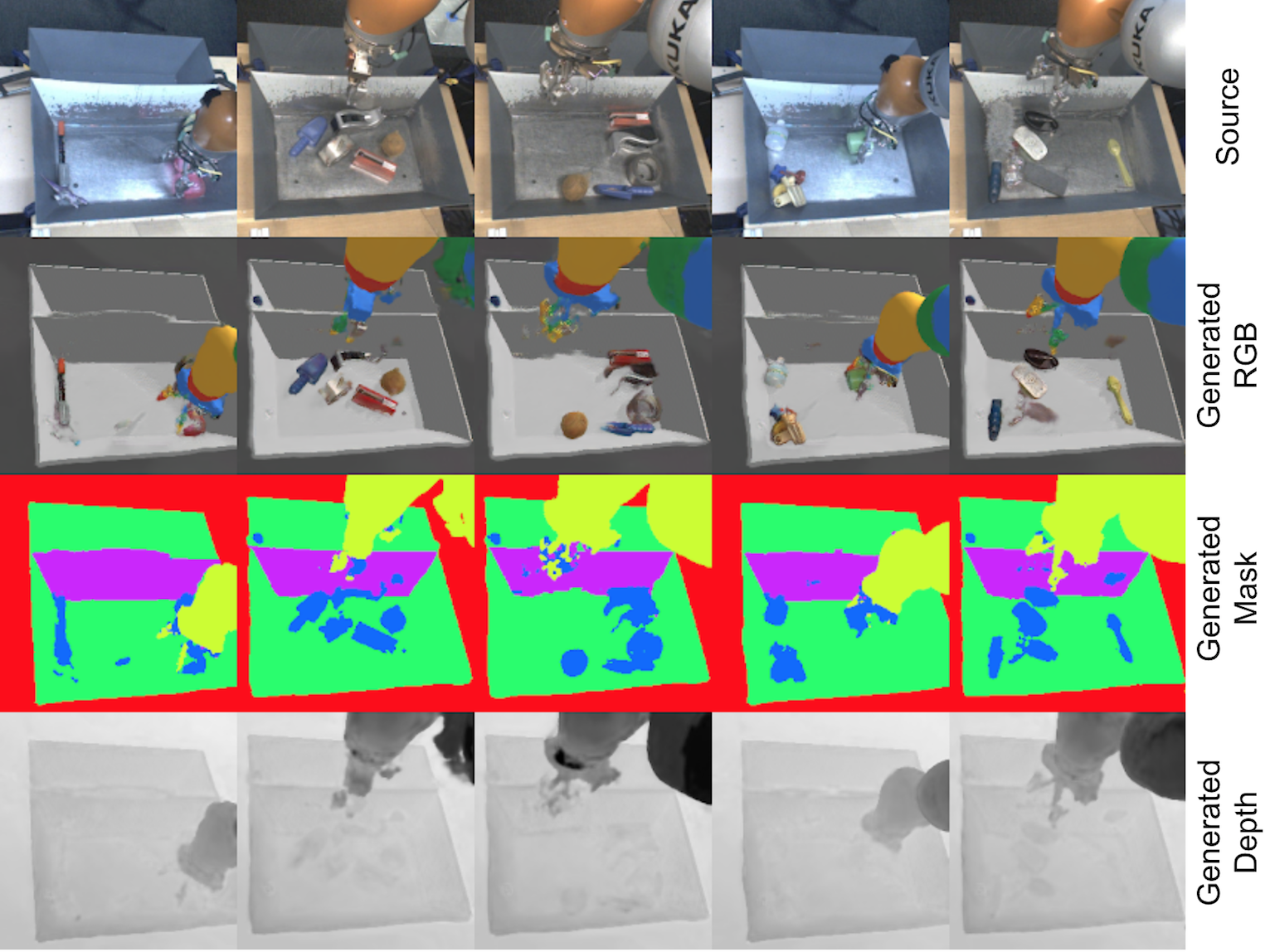}
\caption{Real-to-canonical samples.}
\label{subfig:realtosim}
\end{subfigure}
\caption{Sample outputs of our trained generator $\gen$ when given randomized sim images (\ref{subfig:simtosim}) and real images (\ref{subfig:realtosim}). Note the accuracy of the reconstruction of the canonical images from real-world images in complex and cluttered scenes, along with shadows being re-rendered into the canonical representation. However, also note that randomized-to-canonical adaptation performs a noticeably better reconstruction of the gripper in comparison to the real-to-canonical adaptation. This leads to the failure cases discussed in Section \ref{sec:experiments}. The generated depth and segmentation masks are used as auxiliaries during training of the generator. Further examples can be seen in Figure \ref{fig:appendix_generator_samples} of the Appendix.}
\label{fig:generator_samples}
\end{figure*}

We create the trajectories from which we sample paired snapshots by running training of \qtopt{} in simulation. At the beginning of each episode, the position of the divider in the tray is randomly sampled, and 5 randomly selected objects are dropped into the tray.
Then, at each timestep we freeze the scene, apply a new arbitrary randomization (described below) to capture the randomized observation, reset to and capture an observation of the 
canonical version, and let \qtopt{} proceed.
In our case, observations consist of RGB images, depth, and segmentation masks, labeling each pixel with one of 5 categories: graspable objects, tray, tray divider, robot arm, and background.

The randomization includes applying at each timestep randomly selected textures from a set of over 5,000 images to all models, which includes the tray, graspable objects, arm segments, and floor. Additionally we randomize the position, direction and color of the lighting.
To further increase the diversity of scene configurations beyond those that the normal robot operation during \qtopt{} training gives us, we also slightly randomize the position and size of the arm and tray (sampling from a uniform distribution), applying the same transformation to both the canonical and the randomized scene when creating the snapshot, such that the semantics between the two still match.

One important question is: what should the canonical environment look like? In practice, the canonical environment can be defined in a number of ways. We opt for applying uniform colors to the background, tray and arm, while leaving the textures for the objects from the randomized version in-place, as this preserves the objects' identity and thus opens up the potential for instance-specific grasping in future works. Each link of the arm is colored independently to aid tracking of individual links of the arm. We opt for fixing the light source in the canonical version, requiring the network to learn some aspect of geometry in order to re-render any shadows in the correct shape and direction.

\subsection{RCAN Training Method}
\label{sec:trainingmetHOS}

We aim to learn $\gen(\imagerand) \rightarrow \{\imagefake, \maskfake, \depthfake\}$, which transforms randomized sim images into canonical sim images with matching semantics, with the intuition that the generator will generalize to accept an image from the real world $\imagereal$, and produce a canonical RGB image, segmentation mask, and depth image: $\gen(\imagereal) \rightarrow \{\imagefake, \maskfake, \depthfake\}$. To train the generator, we encourage visual equality between the generated $\imagefake$  and target $\imagecanonical$ through a loss function $l_{eq_\image}$, semantic equality between $\maskcanonical$ and $\maskfake$ through a function $l_{eq_\mask}$, and depth equality between $\depthcanonical$ and $\depthfake$ through a function $l_{eq_\depth}$. Having experimented with L1, L2, and the mean pairwise squared error (MPSE), our solution uses MPSE for $l_{eq_\image}$ which was found to converge faster with no loss in performance~\cite{bousmalis2016domain}, along with the L2 distance for our auxiliary losses $l_{eq_\mask}$ and $l_{eq_\depth}$. This results in the following loss:
\begin{align}
\mathcal{L}_{eq}(\gen) = &\ \E_{(\imagerand, \imagecanonical, \maskcanonical, \depthcanonical)} [ \lambda_{\image} l_{eq_\image}(\gen_{\image}(\imagerand),  \image_{\canonical}) \; + \\ 
& + \lambda_{\mask} l_{eq_\mask}(\gen_{\mask}(\imagerand), \mask_{\canonical}) + \lambda_{\depth} l_{eq_\depth}(\gen_{\depth}(\imagerand), \depth_\canonical) ] , \nonumber
\end{align}
where $\gen_{\image}$, $\gen_{\mask}$, and $\gen_{\depth}$ denotes the image, mask, and depth element of the generator output respectively. In addition, $\lambda_{\image}$, $\lambda_{\mask}$ and $\lambda_{\depth}$ represent the respective weightings.

It is well known that these equality losses can lead to blurry images~\cite{larsen2015autoencoding}, and so we employ a sigmoid-cross entropy generative adversarial (GAN) objective~\cite{goodfellow2014generative} to encourage high-frequency sharpness. Let $\disc(\image)$ be a discriminator that outputs the likelihood that a given image $\image$ is from the canonical domain. With this, the GAN is trained with the following objective:
\begin{equation}
\mathcal{L}_{GAN}(\gen,\disc) = \mathbb{E}_{\image}[\log \disc(\image)] + \mathbb{E}_{\image}[\log (1-\disc(\gen_{\image}(\image))],
\end{equation}
where $\gen_{\image}$ denotes the image element of the generator output. The final objective for the generator then becomes:
\begin{equation}
\hat{\gen} = \arg\min_{\gen} \max_{\disc} \mathcal{L}_{GAN}(\gen,\disc) +  \mathcal{L}_{eq}(\gen) \; .
\label{eq:simonly}
\end{equation}
The generator $\gen$ and discriminator $\disc$ are parameterized by weights of a convolutional neural network; details of which are presented in Appendix \ref{app:rcan}. Qualitative results of our generator can be seen in Figure \ref{fig:generator_samples} and on the project web-page\footnote{\webaddress{}}.

\subsection{Real World Grasping with QT-Opt}
\label{sec:simtosim_qtopt}

We use \qtopt{} for our grasping algorithm, and follow the same state and action definition as Kalashnikov \etal~\cite{Kalashnikov2018}, where the state is defined as $\bs_t = (\image_t, g_{\text{apt},t}, g_{\text{height},t})$ at each timestep $t$, which includes a $472 \times 472$ image $\image_t$ taken from a mounted over-the-shoulder camera overlooking the work space, a binary open/close indicator of gripper aperture $g_{\text{apt},t}$, and the scalar height of the gripper above the bottom of the tray $g_{\text{height},t}$. 

In our case, rather than sending the image directly to the RL algorithm, the image $\image_t$ is instead passed through the generator $\gen$, and the resulting generated image $\imagefake$ is extracted and concatenated, channel-wise, with the original source image $\image_t$. This results in the state $\bs_t = ([\gen(\image_t) + \image_t], g_{\text{apt},t}, g_{\text{height},t})$, where $[\gen(\image_t) + \image_t]$ represents the concatenation. Note that we do not use the generated depth and segmentation masks of $\gen$ as input to \qtopt{} in order to make a fair comparison to Kalashnikov \etal~\cite{Kalashnikov2018}, though these could also be added in practice. The action space of Kalashnikov \etal~\cite{Kalashnikov2018}, which consists of gripper pose displacement and an open/close command, remains unchanged. A summary of the Q-function is shown in Figure \ref{fig:qtopt_network} of the Appendix, and further details of the action space and architecture can be found in Appendix \ref{app:qtopt}. 

In Kalashnikov \etal~\cite{Kalashnikov2018}, the authors take their agent that was trained with $580{,}000$ off-policy real-world grasps, and jointly finetune with an additional 28,000 on-policy grasps. During this joint finetuning process, \qtopt{} asynchronously updates target values, collects real on-policy data, reloads real off-policy (offline) data from past experiences, and then trains the Q-network on both the on and off policy data streams within a distributed optimization framework. In the case of jointly finetuning \acronym{}, we also collect real on-policy data, but rather than using real-world past experiences (which we assume we do not have), we instead leverage the power of our simulation to continuously generate on-policy simulation data, and instead train on these streams of data. During the real world on-policy collection of both approaches, a selection of about $1{,}000$ diverse training objects are used; a sample of which are shown in Figure \ref{fig:real_train_test_objects} of the Appendix. Between 5 and 10 objects are randomly chosen every few hours to be placed in each of the trays until the desired number of joint finetuning grasps are reached.

\begin{table*}
\begin{center}
\begin{tabular}{r|c|c|c||c|c}
\multicolumn{1}{c|}{\qtopt{} Data Source} & \makecell{Offline \\ Real Grasps} & \makecell{Performance \\ In Sim} & \makecell{Performance \\ In Real} & \makecell{Online \\ Real Grasps} & \makecell{Performance \\ In Real} \\
\hline
\hline
\multirow{2}{*}{Real}      & \multirow{2}{*}{580,000}  & \multirow{2}{*}{-} & \multirow{2}{*}{87\%} & +5,000 & 85\%  \\
                                &                   &                      &                      & +28,000             & 96\%          \\
\hline
Canonical Sim                   & 0                 & 99\%                 & 21\%                 & +5,000              & 30\%          \\
Mild Randomization              & 0                 & 98\%                 & 37\%                 & +5,000              & 85\%          \\
Medium Randomization            & 0                 & 98\%                 & 35\%                 & +5,000              & 77\%             \\
\multirow{2}{*}{Heavy Randomization}             & \multirow{2}{*}{0}                 & \multirow{2}{*}{98\%}                 & \multirow{2}{*}{33\%}                 & +5,000              & 85\%             \\
                                &                   &                      &                      & +28,000             & 92\%          \\

\hline
\multirow{2}{*}{\textbf{\acronym{}}}      & \multirow{2}{*}{\textbf{0}}  & \multirow{2}{*}{99\%} & \multirow{2}{*}{\textbf{70\%}} & +5,000 & \textbf{91\%}  \\
                                &                   &                      &                      & +28,000             & \textbf{94\%}     
\end{tabular}
\end{center}
\caption{Average grasp success rate on test objects after 102 grasp attempts on each of the multiple Kuka IIWA robots. The first 4 columns of the table highlight the performance after training on a specified number of real world grasps. Zero grasps implies that all training was done in simulation. The last 2 columns highlight the results of on-policy joint finetuning on a small amount of real-world grasps.}
\label{tbl:results}
\end{table*}

\section{Experiments}
\label{sec:experiments}

Our experimental section aims to answer the following questions: (1) Can we train an agent to grasp arbitrary unseen objects without having seen any real-world images? (2) How does \qtopt{} perform with standard domain randomization, and can our method perform better than this? (3) Does the addition of real-world on-policy training of our method lead to higher grasping performance while still drastically reducing the amount of real-world data required? We answer these questions through a series of rigorous real-world vision-based grasping experiments across multiple Kuka IIWA robots.


\subsection{Evaluation Protocol}

During evaluation, each robot attempts 102 grasps on its own set of 5 to 6 previously unseen test objects (shown in Figure \ref{fig:real_train_test_objects} of the Appendix) which are deposited into each robots' respective tray and remain constant across all evaluations. Each grasp attempt (episode) consists of at most 20 time steps. If after 20 time steps no object has been grasped, the attempt is regarded as a failure. Following a grasp attempt, the object is deposited back into the tray at a random location. Although grasping was done with replacement, in practice, \qtopt{} was not found attempting a grasp on the same object multiple times in a row. All observations come from an over-the-shoulder RGB camera.

\subsection{Results}

We first focus on the first 4 columns of Table \ref{tbl:results}. The first row of this section shows the results of \qtopt{} reported in Kalashnikov \etal~\cite{Kalashnikov2018}; where following 580,000 off-policy real-world grasps, a performance of $87\%$ was achieved. The \textit{Canonical Sim} data source (second row) takes \qtopt{} trained in the canonical simulation environment and then runs this directly in the real-world. The low success rate of $21\%$ shows the existence of the reality gap. The following three rows show the result of training \qtopt{} directly on varying degrees of randomization: mild, medium and heavy. \textbf{Mild randomization} consists of varying tray texture, object texture and color, robot arm color, lighting direction and brightness, and a background image consisting of 6 different images from the view of the real-world camera. \textbf{Medium randomization} adds a diverse mix of background images to the floor. Finally, \textbf{heavy randomization} uses the same scheme used to train \acronym{}, explained in Section \ref{sec:datageneration}.

Surprisingly, an unexpected discovery was that \qtopt{} responds well to heavy domain randomization during training (\ie{} is not destabilized). This is contrary to other RL methods, such as DDPG~\cite{lillicrap2015continuous} and A3C~\cite{mnih2016asynchronous}, where heavy domain randomization has been shown to cause training to fail~\cite{matas2018sim, Zhang2018a}. Although \qtopt{} was able to train stably with randomization, the results show that this does not lead to a successful transfer, achieving between $33\%$ and $37\%$ zero-shot grasping performance, whereas \acronym{} achieves \textbf{70\%}: over \textbf{double} the success in the real world. This success highlights that \acronym{} better utilizes domain randomization to achieve sim-to-real transfer, rather than training a policy directly on domain randomization.

We now focus on the remaining 2 columns, that is, the ability to jointly finetune on a small amount of real-world on-policy grasps. We chose to use $5{,}000$ to represent ``\textit{small}'', which is less than $1\%$ of the $580{,}000$ grasps used in Kalashnikov \etal~\cite{Kalashnikov2018} for the off-policy training and takes only a day to collect, instead of months. To make comparison easier, in addition to reporting the $28{,}000$ on-policy grasps for joint finetuning from \cite{Kalashnikov2018}, we also report the performance after $5{,}000$ grasps. This baseline result of $85\%$ suggest that $5{,}000$ real-world grasps for joint finetuning a system already trained with 580,000 does not improve performance. For the next joint finetuning experiment, we take each of the agents that were trained directly on domain randomization, and jointly finetune them on 5,000 real grasps, achieving between $77\%$ and $85\%$ grasping success. The rapid increase of $\sim50{p{.}p{.}}$ is very surprising, and to the best of our knowledge, no other related works have shown such a dramatic performance increase from pre-training on domain randomization. 

Finally, we look at joint finetuning \acronym{} with 5,000 and 28,000 real grasps, where the real images are adapted by the generator and then both the source and adapted image are passed to the grasping network; in this case, the gradients are only applied to the grasping network and not the generator network. The result of 91\% for $5{,}000$ shows that the improvement over learning directly on domain randomization holds, though for this result the difference is much smaller. What we believe is incredibly encouraging for the robotics community, is that with \textbf{91\% \acronym{} outperforms} a version of \qtopt{} that was trained on 580,000 real-world grasps, while using \textbf{less than 1\% of the data}. Moreover, following joint finetuning with with the same number of online grasps as Kalashnikov \etal~\cite{Kalashnikov2018} (28,000), we are able to achieve an almost equal grasp performance of $94\%$.

\begin{figure}
\begin{center}
\includegraphics[width=0.9\linewidth]{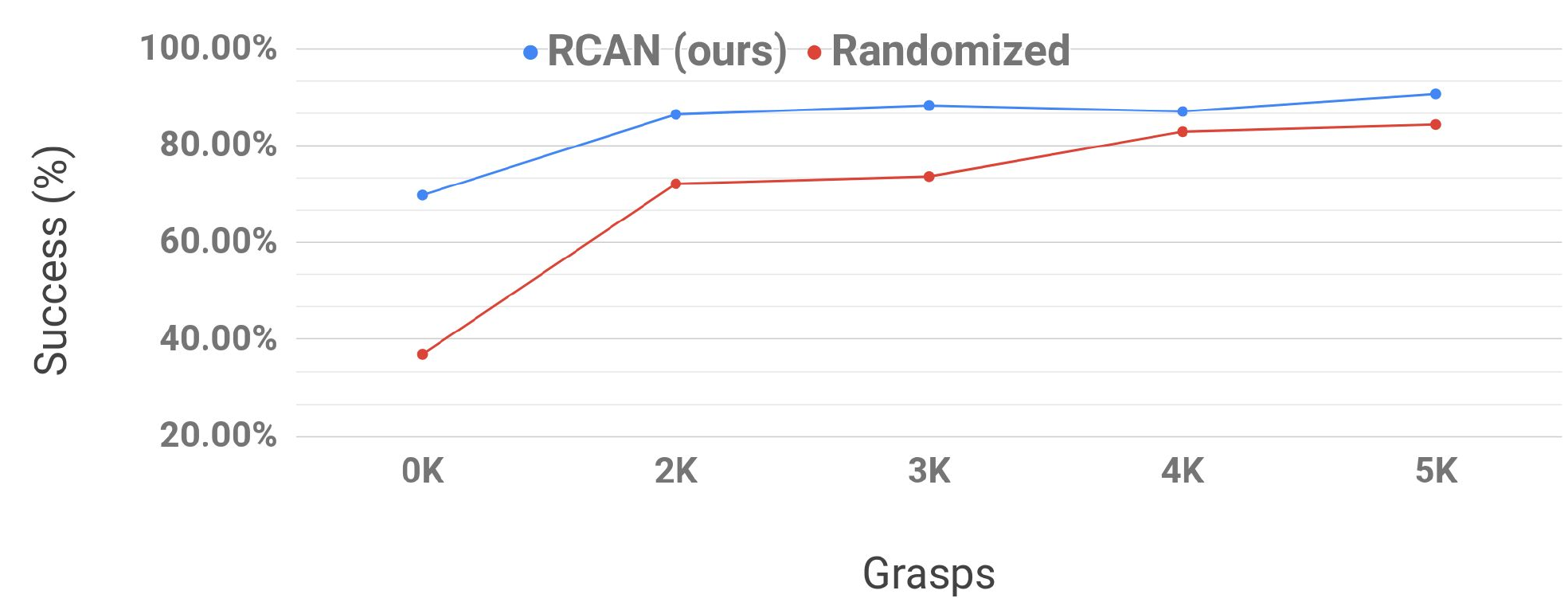}
\vspace{-0.15in}
\end{center}
\caption{A graph showing how the performance of \acronym{} and directly learning a policy on domain randomization varies with the number of real world on-policy grasps.}
\label{fig:grasps_graph}
\end{figure}

In order to understand how performance varies as we progress from 0 to 5,000 on-policy grasps, we repeat the evaluation protocol set above for intermediate checkpoints. We re-evaluate both agents at every 1,000 grasps for both \acronym{} and \textit{Mild Randomization}. The results, presented in Figure \ref{fig:grasps_graph}, show that the majority of the success is gained within the first 2,000 grasps for both approaches. This is encouraging, as we ultimately wish to limit the amount of real-world data that we are reliant on.

\subsection{Failure cases}
\label{sec:failurecases}

A large contributing factor to \qtopt{}s 96\% grasp success, was its ability to perform corrective behaviors, regrasping,  probing motions to ascertain the best grasp, and non-prehensile repositioning of objects. Much of this ability remained with our approach, except for the regrasping ability. This powerful ability allows the policy to detect when there is no object in the closed gripper, and thus, it can decide to re-open it in an attempt to try and re-grasp. Given that our method is not perfect at translating real-world images into simulation ones, artifacts may arise. As objects that we grasp are often small, it can be very difficult for the agent to differentiate between artifacts in the image or if there is indeed an object in the gripper. We observe this to be detrimental to the agents ability to perform regrasping, resulting in only a small amount of regrasps. The main observation from joint finetuning our method with $5{,}000$ real-world grasps, is the re-emergence of the regrasping. We believe that this is contributed by our decision to concatenate the source image to the generated ones, and thus giving the grasping algorithm the option to choose which data source to extract information from for each part of the image as the joint finetuning continues. We hypothesize, that as the number of joint finetuning grasps increase, the network would eventually learn to solely rely on the source (real-world) image, rather than the adapted simulation image. However, we believe that, with a limited amount of labeled real-world data, feeding both the output of \acronym{} as well as the original image to the agent offers the best 
combination of a simplified, yet potentially incomplete adapted view and the complex, but complete original real-world view. 

\subsection{Discussion}
\label{sec:discussion}

A number of questions arise from these results. For example: why does our method perform better than learning a policy directly with domain randomization? We hypothesize that our method allows offloading visual complexity to the generator network, thus simplifying the task for the grasping network and in turn, leading to a higher grasping success. Moreover, having a chosen canonical environment allows us to impose structure on the task which may be beneficial for training the grasping network.. Despite our method achieving over double the zero-shot performance in the real world in comparison to domain randomization, with $5{,}000$ additional real-world grasps, the performance of direct domain randomization also achieves a surprisingly high performance. This leads us to the hypothesis that learning a policy directly on domain randomization can act as a very powerful pre-training regime, where the network is forced to learn a very general feature extractor that can be easily jointly finetuned to a new environment. Having said that, our method outperforms this and has the added benefit of giving us an interpretable output for sim-to-real transfer.

Another question for future work would be: is there a way to better utilize the data collected during the $5{,}000$ on-policy grasps? Given this real-world data, it is now possible to consider fusing ideas from other transfer methods that require some real-world data, such as PixelDA~\cite{bousmalis2016domain}.

\section{Conclusion}

We have presented \methodname{} (\acronym{}), a sim-to-real method that learns to translate randomized simulation images into a canonical representation, which in turn allows for real-world images to also be translated to this canonical representation. Given that our grasping algorithm (\qtopt{}) is trained in this canonical environment, it is possible to run policies trained in simulation in the real world. We show that this approach is superior to the common domain randomization approach, and argue that it is a much more meaningful use of domain randomization. This general style of transfer has applications beyond just grasping, and can be used in other settings where real world data is expensive to collect, for example, producing segmentation masks for self-driving cars. For future work, we wish to explore further ways of introducing unlabelled real-world data in order to improve the real-to-canonical translation. Moreover, we are interested in exploring the effect of using the auxiliary outputs as additional inputs to the grasping network.

\section*{Acknowledgments}
We would like to give special thanks to Ivonne Fajardo, Peter Pastor, I\~{n}aki Gonzalo and Benjamin Swanson for overseeing the robot operations, Yunfei Bai for discussion on PyBullet, and Serkan Cabi for valuable comments on the paper.

{\small
\bibliographystyle{ieee}
\bibliography{egbib}
}

\clearpage

\appendix

\section{RCAN Architecture}
\label{app:rcan}

\begin{figure}
\includegraphics[width=1.0\linewidth]{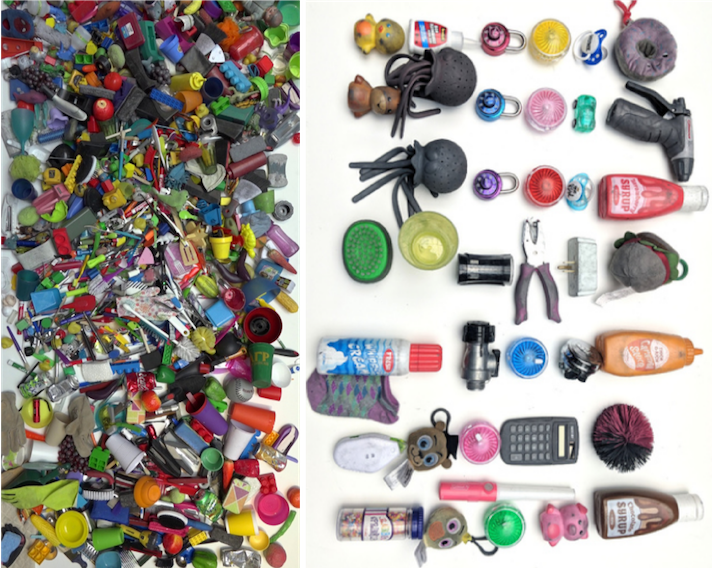}
\caption{Real-world grasping objects that range greatly in size and appearance. \textit{Left}: about $1000$ visually and physically diverse training objects used for joint finetuning. \textit{Right}: the unseen test objects.}
\label{fig:real_train_test_objects}
\end{figure} 

\begin{figure*}
\includegraphics[width=1.0\linewidth]{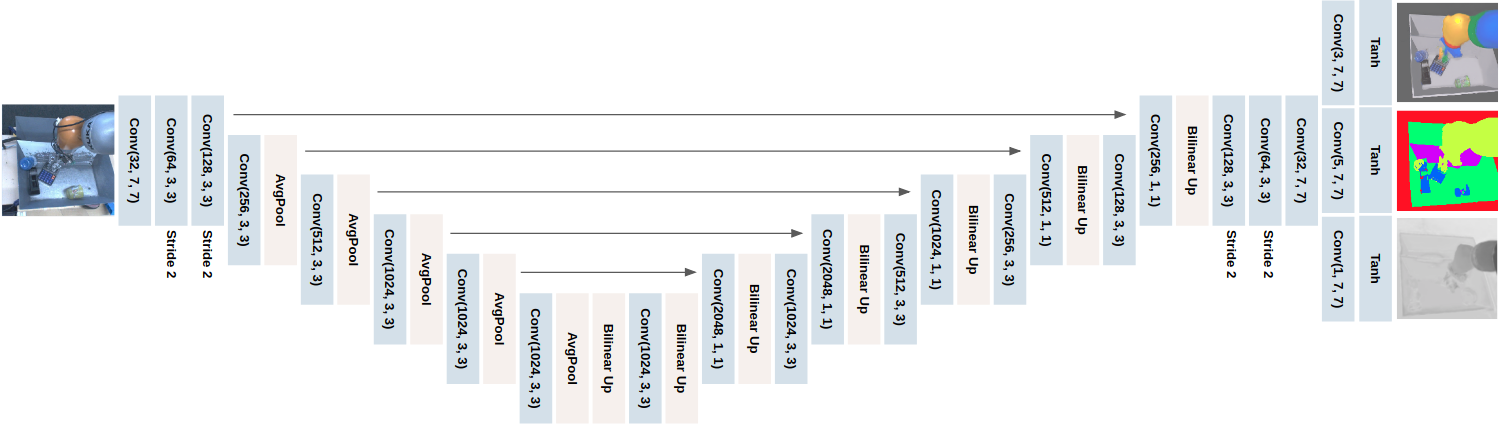}
\caption{Network architecture of the generator function $\gen$. An RGB image from the source domain (either from the randomized domain or real-world domain) is processed via a U-Net style architecture~\cite{ronneberger2015u} to produce a generated RGB image $\imagefake$, and auxiliaries that includes a segmentation mask $\maskfake$ and depth image $\depthfake$. These auxiliaries forces the generator to extract semantic and depth information about the scene and encode them in the intermediate latent representation, which is then available during the generation of the output image.}
\label{fig:unet_arch}
\end{figure*}

The generator $\gen$ is parameterized by weights of a convolutional neural network, summarized in Figure \ref{fig:unet_arch}, and follows a U-Net style architecture~\cite{ronneberger2015u} with downsampling performed via $3 \times 3$ convolutions with stride 2 for the first 2 layers, and average pooling with $3 \times 3$ convolution of stride 1 for the remaining layers. Upsampling was performed via bilinear upsampling, followed by a $3 \times 3$ convolutions of stride 1, and skip connections were fused back into the network via channel-wise concatenation, followed by a $1 \times 1$ convolution. All layers were followed by instance normalization~\cite{ulyanov1607instance} and ReLU non-linearities. The discriminator $\disc$ is also parameterized by weights of a convolutional neural network with 2 layers of 32, $3 \times 3$ filters, followed by a layer of 64, $3 \times 3$ filters, and finally a layer of 128, $3 \times 3$ filters. The network follows a multi-scale patch-based design~\cite{Bousmalis2018}, where 3 scales of $472 \times 472$, $236 \times 236$, and $118 \times 118$, are used to produce domain estimates for all patches which are then combined to compute the joint discriminator loss. The weightings $\lambda_{\image}$, $\lambda_{\mask}$ and $\lambda_{\depth}$ in Equation 2 were all set to $1$.

\section{QT-Opt Architecture}
\label{app:qtopt}

\begin{figure}
\includegraphics[width=1.0\linewidth]{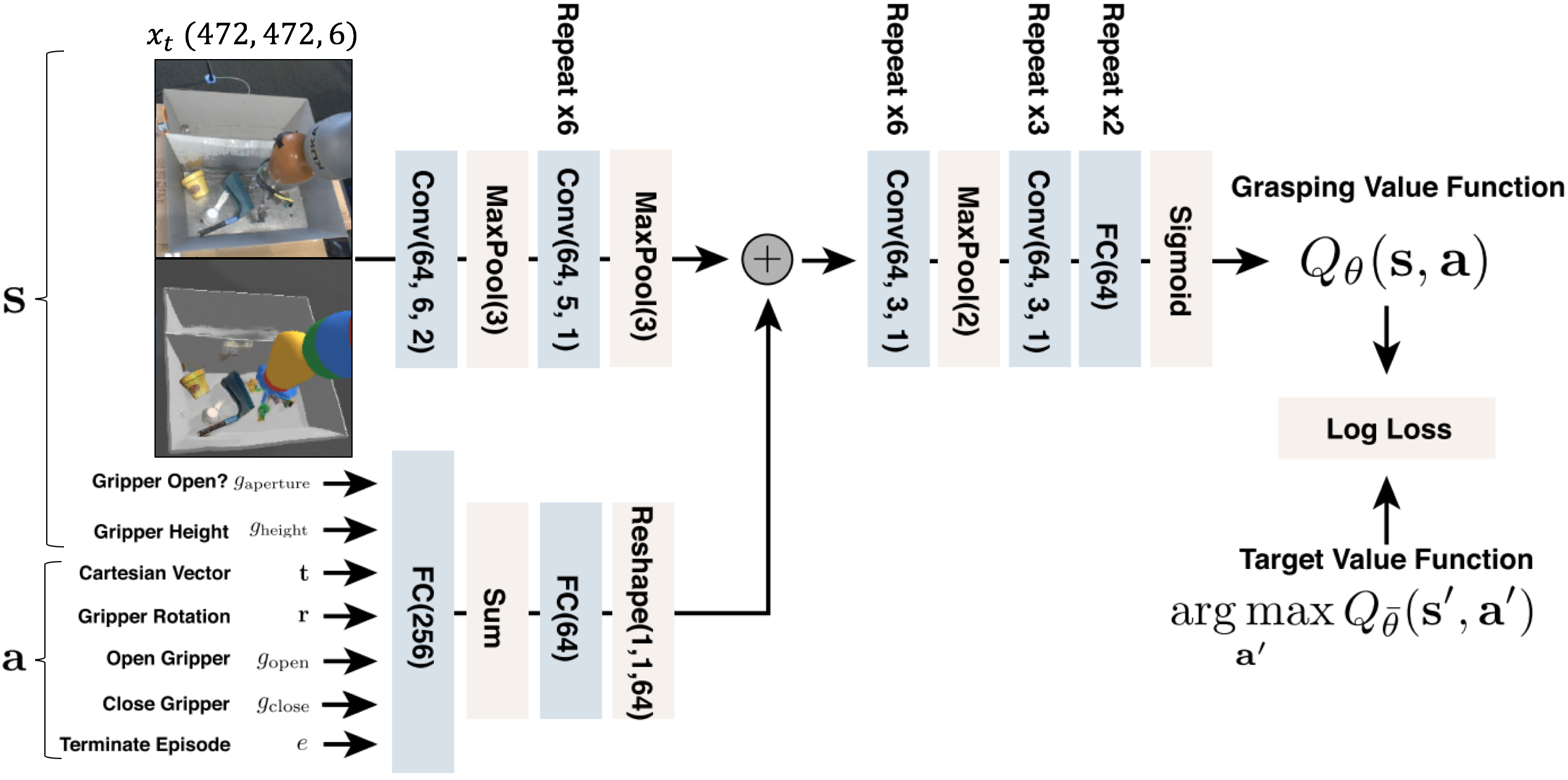}
\caption{The Q-function of the grasping algorithm. The source image $\image$ (either from the randomized domain or real-world domain) and generated canonical image $\imagefake$ are concatenated (channel-wise) and processed by a convolutional neural network (and fused with action and state variables) to produce a scalar representing the Q value $Q_\theta(s,a)$.}
\label{fig:qtopt_network}
\end{figure}

The action space of \cite{Kalashnikov2018}, which consists of gripper pose displacement and an open/close command, remains unchanged in our paper, and is defined as $\ba_t = (\mathbf{t}_t, \mathbf{r}_t, g_{\text{close},t}, g_{\text{open},t}, e_t)$, containing Cartesian translation $\mathbf{t}_t \in \mathbb{R}^3$, sine-cosine rotation encoding $\mathbf{r}_t \in \mathbb{R}^2$, a one-hot vector gripper open/close command $[g_{\text{close},t}, g_{\text{open},t}] \in \{0, 1\}^2$, and a learned stopping criterion $e_t$. The reward function is sparse, consisting of a reward of $1$ following a successful grasp, or $0$ for an unsuccessful grasp, and $-0.05$ on all other transitions. Summarized in Figure \ref{fig:qtopt_network}, the Q-function follows the same architecture as \cite{Kalashnikov2018} (originally inspired by \cite{Levine2016}).

Rather than a single RGB image input, our network takes in a 6 channel image, consisting of channel-wise concatenation of the source image $\image$ (either from the randomized domain or real-world domain) and generated image $\imagefake$. Features are extracted from these images via 7 convolutional layers and then merged with a transformed action and state vector (which have passed through 2 fully-connected layers) via element-wise addition. The merged streams are then processed by a further 9 convolution layers and 2 fully-connected layers, resulting in a scalar output representing the Q value $Q_\theta(s,a)$. Each layer, excluding the final, uses batch normalization~\cite{ioffe2015batch} and ReLU non-linearities. A summary of the architecture can be seen in Figure \ref{fig:qtopt_network}.

\begin{figure*}
\centering
\begin{subfigure}[b]{1.0\linewidth}
\includegraphics[width=\textwidth]{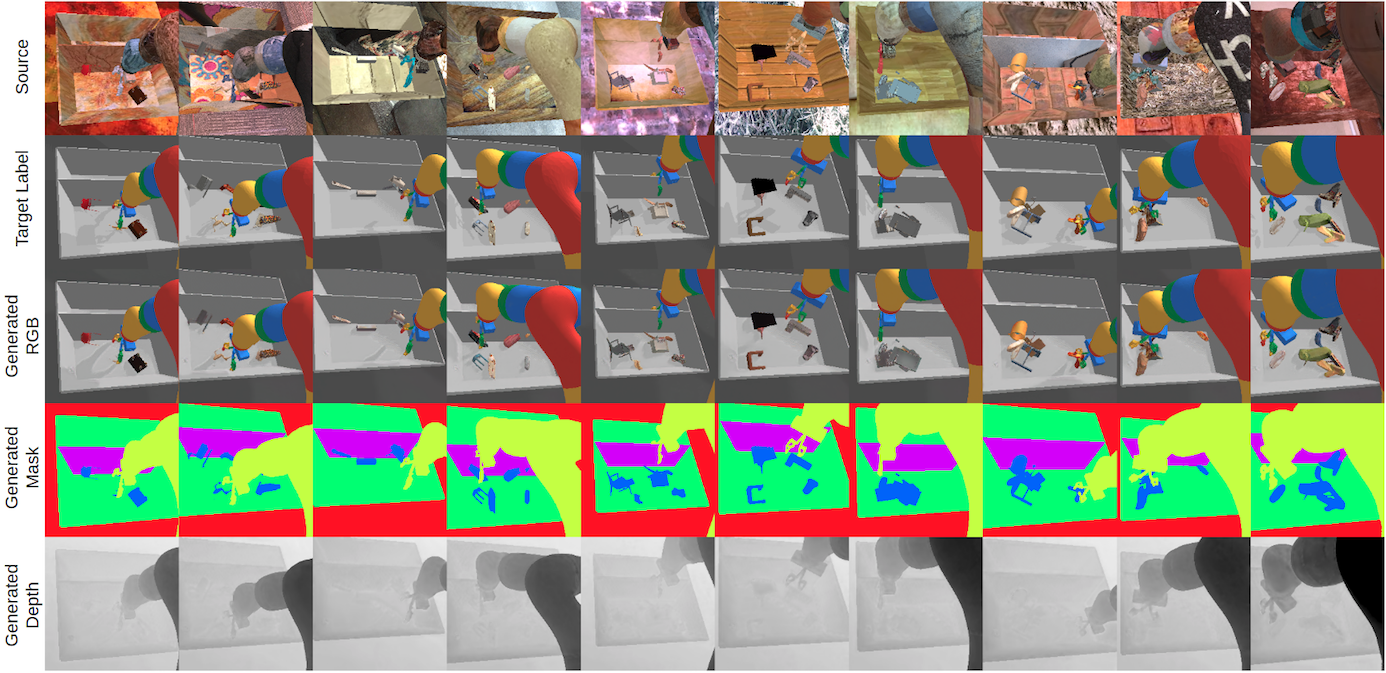}
\caption{Randomized-to-canonical samples.}
\label{subfig:appendix_simtosim}
\end{subfigure}
\begin{subfigure}[b]{1.0\linewidth}
\includegraphics[width=\linewidth]{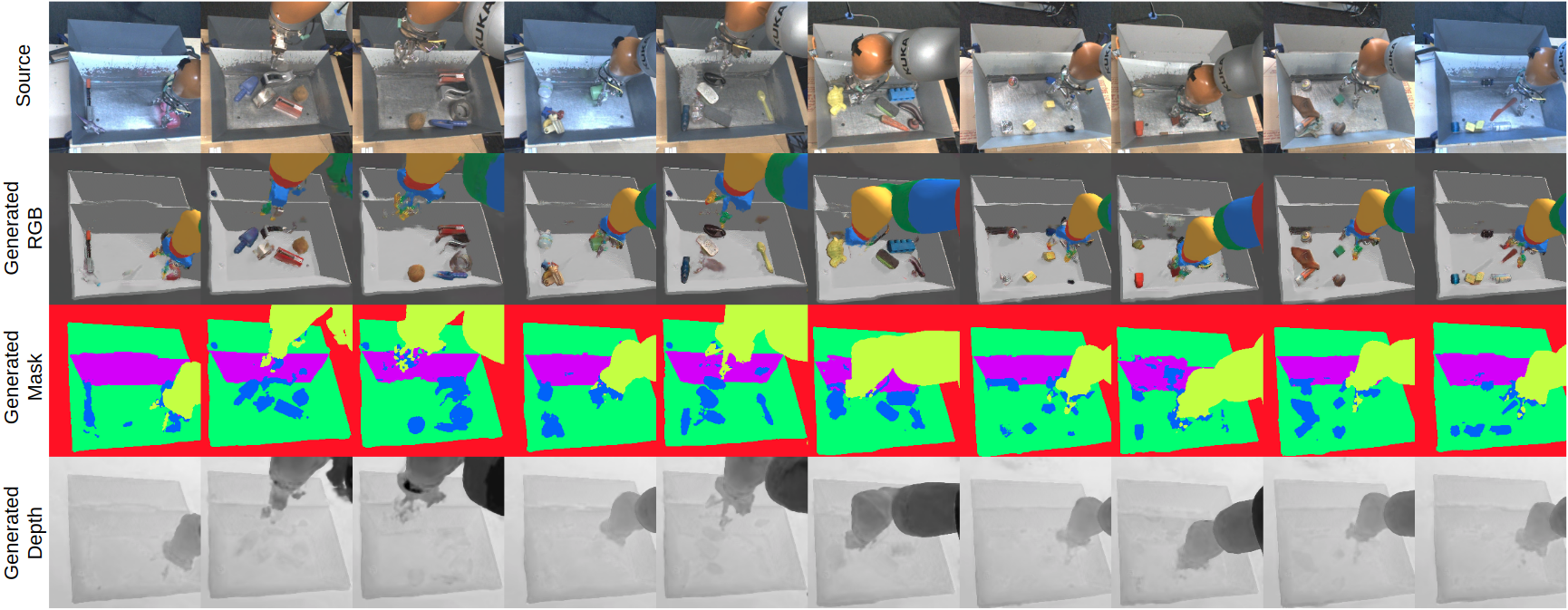}
\caption{Real-to-canonical samples.}
\label{subfig:appendix_realtosim}
\end{subfigure}
\caption{Additional sample outputs of our trained generator $\gen$ when given randomized sim images (\ref{subfig:appendix_simtosim}) and real images (\ref{subfig:appendix_realtosim}).}
\label{fig:appendix_generator_samples}
\end{figure*}

\end{document}